%% file: main.tex
\documentclass[conference]{IEEEtran}
\IEEEoverridecommandlockouts
\usepackage{epsfig}
\usepackage{amsmath}
\usepackage{amssymb}
\usepackage{color}
\usepackage{url}
\usepackage[noabbrev,capitalize]{cleveref}
\usepackage[utf8]{inputenc} % allow utf-8 input
\usepackage[T1]{fontenc}    % use 8-bit T1 fonts
\usepackage{url}            % simple URL typesetting
\usepackage{booktabs}       % professional-quality tables
\usepackage{amsfonts}       % blackboard math symbols
\usepackage{amsmath}
\usepackage{algpseudocode}
\usepackage{algorithm}
\usepackage{nicefrac}       % compact symbols for 1/2, etc.
\usepackage{microtype}      % microtypography
\usepackage{xcolor}         % colors
\usepackage{colortbl}
\usepackage{graphicx}
\usepackage{tabularx}
\usepackage{adjustbox}
\usepackage{tabularx}
\usepackage{nicematrix}
\usepackage{siunitx}
\usepackage{makecell}
\usepackage{authblk}
\makeatletter
\renewcommand\AB@affilsepx{, \protect\Affilfont}
\makeatother

\newcommand{\qlatentat}[1]{\mathbf{\hat{y}}_{#1}}

\newcommand{\na}{%
  \settowidth{\dimen0}{\num{-0}}%
  \settowidth{\dimen2}{\num{-.}}%
  \makebox[\dimen2][l]{\hspace*{\dimen0}\makebox[0pt]{ -}}%
}

\def\BibTeX{{\rm B\kern-.05em{\sc i\kern-.025em b}\kern-.08em
    T\kern-.1667em\lower.7ex\hbox{E}\kern-.125emX}}
\begin{document}

\title{DCVC-MB: Neural B-Frame Video Compression using State Space Models}

\author[1]{Arjun Arora$^{*\dagger}$\thanks{$\dagger$ Work done while at Dolby. Currently at Voia}}
\author[1]{Calvin-Khang Ta$^{*\ddagger}$\thanks{$\ddagger$ Corresponding Author (czta@dolby.com)}}
\author[3]{Carlos Restrepo-Galeano$^{*\|}$\thanks{$\|$ Work done during internship at Dolby}}
\author[2]{Kruthi Murali}
\author[2]{Naga Akhil E S}
\author[2]{\authorcr Arunkumar Mohananchettiar}
\author[2]{Jay Shingala}
\author[1]{Tong Shao}
\author[1]{Peng Yin}
\author[1]{Sean McCarthy}
\affil[1]{Dolby Laboratories}
\affil[2]{Ittiam Systems}
\affil[3]{University of Delaware}

\maketitle
\renewcommand{\thefootnote}{\fnsymbol{footnote}}
\footnotetext[1]{Equal contribution.}

\input{sections/0_abstract.tex}
\input{sections/1_introduction.tex}

\input{sections/2_related_works.tex}
\input{sections/3_methodology.tex}

\input{sections/4_experiments.tex}

\input{sections/ablation.tex}

\input{sections/5_conclusion.tex}

\bibliographystyle{IEEEbib}
\bibliography{refs}

\end{document}

\title{DCVC-MB: Neural B-Frame Video Compression using State Space Models Appendix}

\author[1]{Arjun Arora$^{*\dagger}$\thanks{$\dagger$ Work done while at Dolby. Currently at Voia}}
\author[1]{Calvin-Khang Ta$^{*\ddagger}$\thanks{$\ddagger$ Corresponding Author (czta@dolby.com)}}
\author[3]{Carlos Restrepo-Galeano$^{*\|}$\thanks{$\|$ Work done during internship at Dolby}}
\author[2]{Kruthi Murali}
\author[2]{Naga Akhil E S}
\author[2]{\authorcr Arunkumar Mohananchettiar}
\author[2]{Jay Shingala}
\author[1]{Tong Shao}
\author[1]{Peng Yin}
\author[1]{Sean McCarthy}
\affil[1]{Dolby Laboratories}
\affil[2]{Ittiam Systems}
\affil[3]{University of Delaware}

\maketitle
\renewcommand{\thefootnote}{\fnsymbol{footnote}}
\footnotetext[1]{Equal contribution.}
% \input{sections/0_abstract.tex}
% \begin{IEEEkeywords}
% Deep Learning, Video Compression, State Space Models
% \end{IEEEkeywords}
% \input{sections/1_introduction.tex}
% \input{sections/2_related_works.tex}
% \input{sections/3_methodology.tex}
% \input{sections/4_experiments.tex}
% \input{sections/ablation.tex}
% \input{sections/5_conclusion.tex}
\input{sections/x_appendix.tex}

\clearpage
\bibliographystyle{IEEEbib}
\bibliography{refs}

% \vspace{12pt}
% \color{red}
% IEEE conference templates contain guidance text for composing and formatting conference papers. Please ensure that all template text is removed from your conference paper prior to submission to the conference. Failure to remove the template text from your paper may result in your paper not being published.

%% file: sections/0_abstract.tex
\begin{abstract}
In this paper we propose DCVC-Mamba (DCVC-MB), a neural video codec framework for B-frame coding. Our approach incorporates an IBP frame strategy for low-delay B-frame coding, a spatio-temporal fusion model based on state-space models for bidirectional temporal prediction, and an entropy-aware skipping mechanism that selectively omits coding certain latents to reduce entropy coding times. In addition to our model contributions we also implement two inference-time strategies that enhance compression performance. Experimental evaluation shows that DCVC-MB compares favorably to existing NVCs and traditional codecs. The method demonstrates BD-rate reductions of up to $8.98\%$ on average compared to prior neural video codecs, and improvements of up to $30.45\%$ and $1.81\%$ over the VTM-19.0-LDP and VTM-19.0-RA(Inter-GoP=16) benchmarks, respectively, contributing to advances in neural video compression.

\end{abstract}

%% file: sections/1_introduction.tex
\section{Introduction}

\begin{figure}[t!]
    \centering
    \includegraphics[width=.9\linewidth]{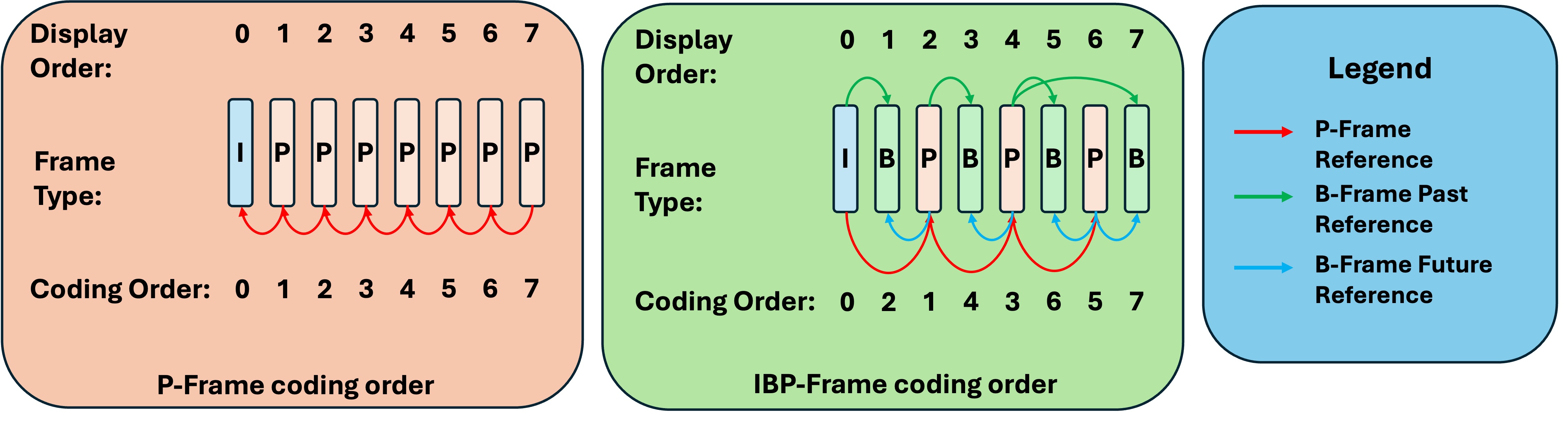}
    \caption{Diagrams showing $a)$ LDP coding using P-frame ordering and $b)$ RA coding using IBP-frame ordering, both in a Closed GoP configuration. Arrows indicate reference relationships.}
    \label{fig:Frame_Ordering}
\end{figure}

The growing demand for efficient video compression has driven innovation in both traditional codecs (AVC~\cite{itu2003264}, HEVC~\cite{6316136}, VVC~\cite{9503377}) and neural video codecs (NVCs)~\cite{li2021deep}. However, leading NVCs~\cite{li2023neural} predominantly use P-frame structures (\cref{fig:Frame_Ordering}), referencing only past frames and forgoing the compression gains from bidirectional prediction that B-frames provide in traditional codecs.

While B-frames provide substantial compression gains in traditional codecs, neural video codecs (NVCs) have struggled to effectively 
leverage them. The core challenge lies in efficiently fusing bidirectional temporal information. Previous attempts \cite{10891533} either underperform P-frame methods or introduce 
prohibitive computational complexity through transformer-based architectures that scale quadratically with resolution.

We address this through DCVC-MB, which introduces a novel bidirectional feature 
fusion architecture based on state-space models (Mamba). Unlike transformers, 
Mamba achieves linear complexity O(N) while effectively capturing long-range 
spatial/temporal dependencies in both directions. This enables practical B-frame coding 
at high resolutions (720p, 1080p) where transformer-based approaches exhaust VRAM. In this paper we make the following contributions:

\begin{itemize}
    \item A base neural compression model to perform Random Access video coding using IBP frame ordering.
    \item A novel feature fusion architecture leveraging state-space models that exploits the temporal redundancy of previously decoded frames.
    \item An adaptive entropy based latent-skipping technique for faster entropy coding.
    \item Two inference-time optimizations, open GoP coding and Bidirectional Coding to enhance the compression performance of our models.

    \item Strong experimental results (Table \ref{tab:combined_bd_rates}): \textbf{30.45\%}/\textbf{1.81\%} BD-Rate savings over VTM-19.0-LDP (Intra-period = -1/32), \textbf{20.19\%} over the much stronger VTM-19.0-RA (Inter-GoP = 16), and \textbf{8.98\%}/\textbf{9.21\%} over DCVC-FM \cite{li2024neural} across both settings.
\end{itemize}

%% file: sections/2_related_works.tex
\section{Related Works}

\begin{figure*}[ht!]
    \centering
    \includegraphics[width=.70\textwidth]{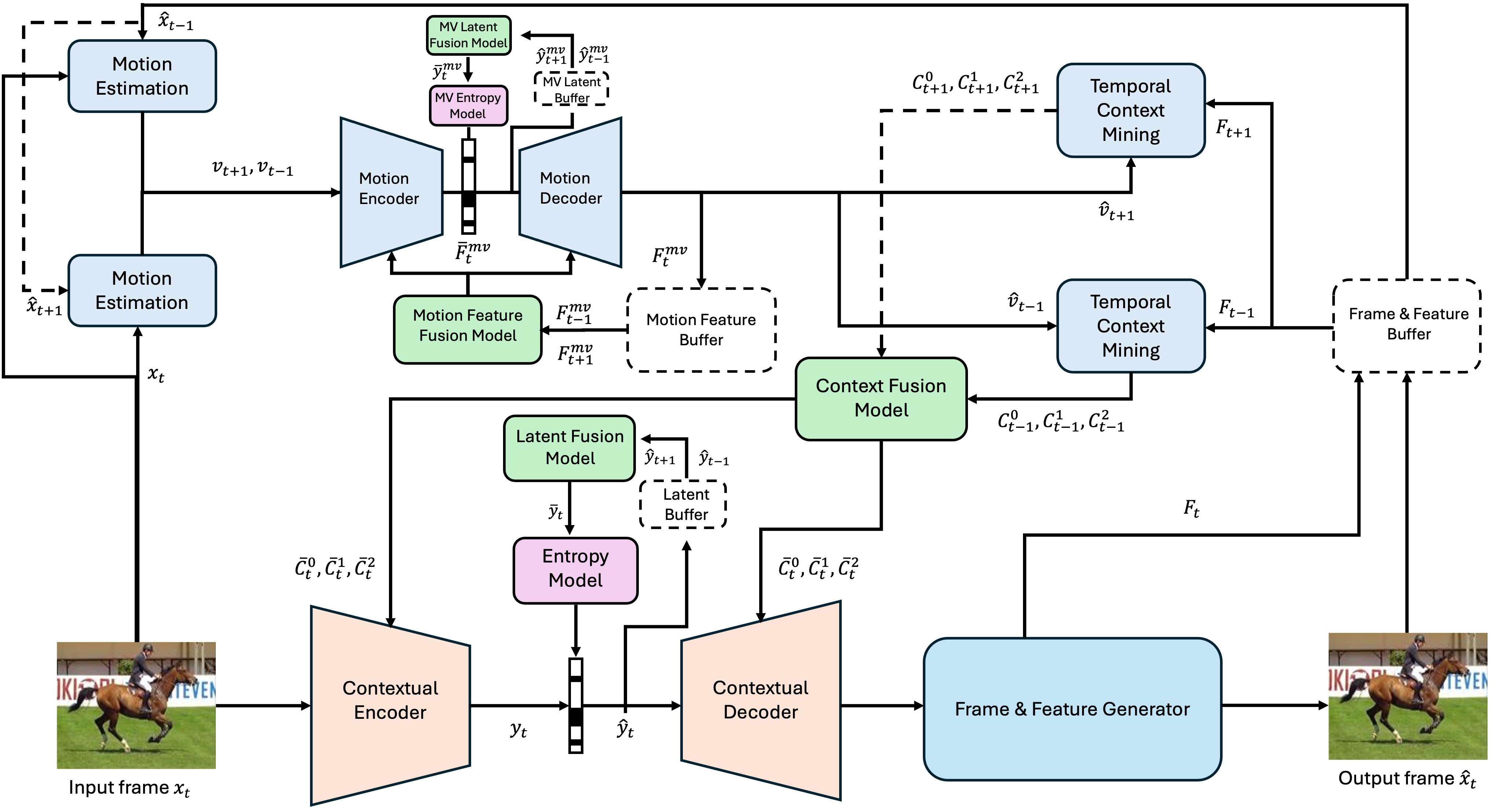}
    \caption{Overview of DCVC-MB architecture, built upon DCVC-DC~\cite{li2023neural}. Unlike prior DCVC models with a single decoded picture buffer (dpb), DCVC-MB maintains two dpbs from the past frame $X_{t-1}$ and future frame $X_{t+1}$, fused via the contextual fusion modules (green) to condition both the encoder and entropy model for B-frame $X_t$.}
    \label{fig:B-Frame Diagram}
\end{figure*}

Neural video codecs build upon image compression techniques \cite{balle2016end,balle2018variational,toderici2017full,bai2022towards,zhu2022transformer}, 
evolving from residual-based methods \cite{ScaleSpaceFlow,lu2019dvc,shi2022alphavc,zhu2022transformer} to context-based approaches 
\cite{ho2022canf,li2021deep,li2022hybrid,li2023neural,li2024neural,wang2023EVC}. However, these predominantly use P-frame structures, 
limiting compression efficiency. Recent B-frame NVCs \cite{kirillov2024hierarchicalbframevideocoding,10891533,yang2024ucvcunifiedcontextualvideo,yilmaz2024motionadaptiveinferenceflexiblelearned} employ 
hierarchical encoding but underperform P-frame methods due to ineffective 
bidirectional fusion and computational complexity. Tang et al. \cite{tang2025neural} propose context modulation for neural video compression but provide no public checkpoints, precluding direct comparison.

Mamba \cite{gu2024mambalineartimesequencemodeling}, a selective state-space architecture, offers O(N) complexity 
versus transformers' O(N²), enabling high-resolution deployment \cite{li2024videomamba,liu2024vmambavisualstatespace}. 
While concurrent work applies Mamba to video compression \cite{qin2024mambavc}, these 
efforts remain limited to P-frame coding. DCVC-MB leverages Mamba for 
efficient bidirectional fusion in B-frame coding, achieving scalability at 1080p where transformer-based approaches fail.

We introduce DCVC-MB, which adopts a simplified RA coding strategy using IBP frame ordering—avoiding hierarchical complexity while retaining bidirectional benefits. We leverage Mamba architectures~\cite{gu2024mambalineartimesequencemodeling,li2024videomamba} for efficient reference feature fusion, exploiting temporal redundancy more effectively.

%% file: sections/3_methodology.tex
\section{Proposed Method}

Our DCVC-MB architecture introduces a fundamental change to the neural video compression framework by enabling simultaneous access to bidirectional temporal information. Unlike prior DCVC models~\cite{li2021deep,li2022hybrid,li2023neural,li2024neural} that rely on a single Decoded Picture Buffer ($dpb$) from the previous frame, our architecture (\cref{fig:B-Frame Diagram}) leverages two $dpb$s concurrently—one from the past and one from the future. This bidirectional design enables fusion of reference features and frames, unlocking greater compression efficiency compared to unidirectional approaches. We maintain the I and P-frame model from~\cite{li2023neural} for compatibility, our key innovation being the B-frame processing that exploits temporal redundancy in both directions.

\subsection{Encode/Decode}

We encode frame $x_t$ into latent $y_t$ using a contextual encoder conditioned on fused multi-scale features $\bar{C}^0_t, \bar{C}^1_t, \bar{C}^2_t$ from adjacent frames $C^i_{t-1}$ and $C^i_{t+1}$. After quantization and entropy coding, the decoded latent $\hat{y}_t$ is processed through a contextual decoder with the fused pyramid to generate reconstructed frame $\hat{x}_t$ and feature $F_t$. Motion vectors $v_{t-1}, v_{t+1}$ follow a similar encoding/decoding process using dedicated modules, but with a single fused motion feature $\bar{F}^{mv}_t$ instead of a full pyramid.

\subsection{Compression}
To compress the latent representations ($y_t$ or $y^{mv}_t$), we use a four-stage entropy model following \cite{li2023neural}. The key difference is that we fuse two decoded reference latents (e.g., $\hat{y}_{t-1}$ and $\hat{y}_{t+1}$ or $\hat{y}^{mv}_{t-1}$ and $\hat{y}^{mv}_{t+1}$ ) to form a single fused latent ($\bar{y}_t$ or $\bar{y}^{mv}_t$), which conditions the entropy model. This model predicts a probability distribution used for entropy coding.

\subsection{Fusion Model}
\begin{figure*}[!t]
    \centering
    \includegraphics[width=.55\linewidth]{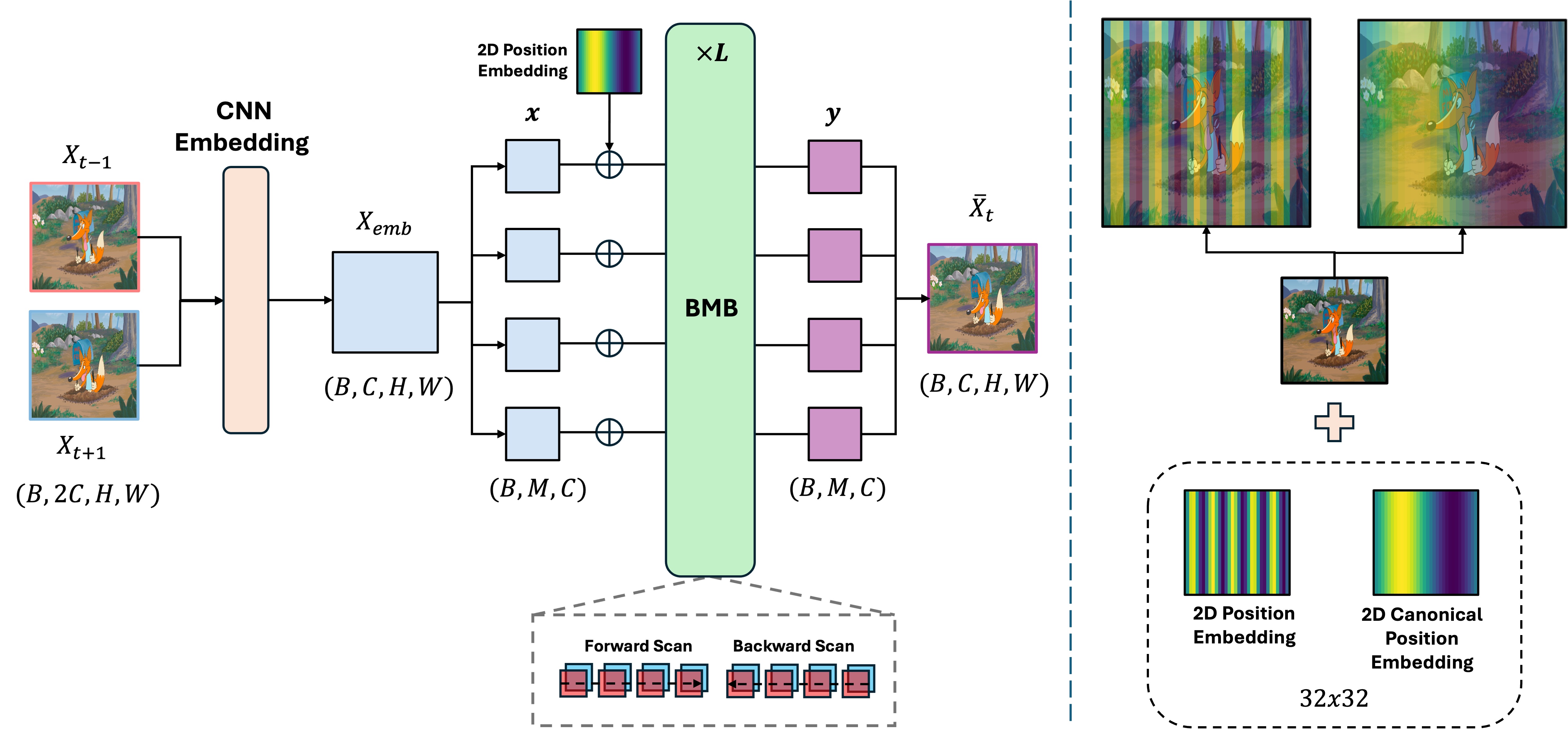}
    \caption{Bidirectional Mamba Fusion Model architecture. Left: Frames $X_{t-1}$ and $X_{t+1}$ are fused into output $\bar{X_t}$ via the Bidirectional Mamba Block (BMB), enabling bidirectional temporal dependency learning. Right: Canonical vs. standard 2D position embeddings. By preserving the spatial harmonics, our canonical embeddings generalize across resolutions.}

    \label{fig:Mamba bidirectional block}

\vspace{-1em}
\end{figure*}

For feature fusion (\cref{fig:B-Frame Diagram}, green blocks), we use the same Bidirectional Mamba Fusion architecture for each fusion model (\cref{fig:Mamba bidirectional block}). The only variation between models lies in the number of input channels, matched to the features being fused. Given inputs $X_{t-1}, X_{t+1} \in \mathbb{R}^{(B,C,H,W)}$, we first concatenate them along the channel dimension to obtain $X_{\text{cat}} \in \mathbb{R}^{(B,2C,H,W)}$. A 2D CNN then embeds this into $X_{\text{emb}} \in \mathbb{R}^{(B,C,H,W)}$, which is tokenized $\textbf{x} \in \mathbb{R}^{(B,M,C)}$ with $M = H \times W$. We add a 2D positional embedding $\textbf{p}_{\textbf{ce}}$
% (see \cref{Canonical Position Emb} ) 
to $\textbf{x}$ indicate the original 2D spatial position of each token.    
% \textbf{x} = \textbf{x} + \textbf{p}_{\textbf{ce}}
% \[ \textbf{x} = \textbf{x} + \textbf{p}_{\textbf{ce}}\]
This sequence is processed by $L = 2$ layers of the Bidirectional Mamba Block (BMB), producing output embeddings $\textbf{y} \in \mathbb{R}^{(B,M,C)}$. Finally, we reshape $\textbf{y}$ back to the spatial format $\bar{X}_t \in \mathbb{R}^{(B,C,H,W)}$ as the fused output.

\subsection{Bidirectional Mamba}
To compute $\textbf{y}$, we have to pass $\textbf{x}$ through the internal State Space Model (SSM). Formally, the Mamba SSM is defined via the following ODE \cite{gu2024mambalineartimesequencemodeling} as 

\begin{align}
    h'(m) = \mathbf{A}h(m) + \mathbf{B}x(m), y(m) = \mathbf{C}h(m)
\end{align}
where $x(m),y(m) \in \mathbb{R}^{C}$ represent the continuous versions of our input $\textbf{x}$ and output $\textbf{y}$ at the $m\text{th}$ token. $h(m) \in \mathbb{R}^N$ is our hidden state, computed using $\mathbf{A} \in \mathbb{R}^{N\times N}$ which represents the parameterized dynamics of the system while $\mathbf{B}, \mathbf{C} \in \mathbb{R}^{N\times 1}$ are our input-dependent projection matrices. However, to process our discrete input sequence, $\textbf{x}_m$, we must discretize the ODE above using a timescale parameter, $\boldsymbol{\Delta}$. Using the zero order hold rule results in $\mathbf{\bar{A}} = exp(\boldsymbol{\Delta}\textbf{A})$ and 
$\mathbf{\bar{B}} = (\boldsymbol{\Delta}\textbf{A})^{-1}(exp(\boldsymbol{\Delta}\textbf{A}) - \textbf{I}) \cdot \boldsymbol{\Delta}\textbf{B}$. This allows us to then represent our final discrete ODE as

$\textbf{h}_m  = \mathbf{\bar{A}}\textbf{h}_{m-1} + \mathbf{\bar{B}}\textbf{x}_m$ and 
$\textbf{y}_m = \mathbf{C}\textbf{h}_{m}$.

This set of equations describe how the SSM can model the dynamics of a system in the forward row-major scan. This assumes that token dependency only operates in the forward direction. However, our visual tokens may have much more complex dynamics that are not captured with only this direction. To solve this problem, we adopt the Spatial-First Bidirectional method described in \cite{li2024videomamba}. We create a second set of parameter matrices, $\textbf{A}_{-m},\textbf{B}_{-m},\textbf{C}_{-m}$ to operate on the backward row-major scan direction, $\textbf{x}_{-m}$, to produce the output sequence $\textbf{y}_{-m}$. $\textbf{y}_{-m}$ is then 
un-reversed and added to $\textbf{y}_{m}$ to compute the final output of the BMB.

\subsection{Canonical Position Embedding}

In most Vision Transformers (ViTs) and Mamba vision papers, it is common to encode spatial position information using a spatial patch's pixel location \cite{16x16Words, AttentionisAllYouNeed}, or relative position to other image patches \cite{Conditional2DPositionEncoding,Explicit2DPositionEncoding}. Because we train on the Vimeo90k dataset \cite{Vimeo90k}, there is a resolution mismatch between training ($256 \times 256$) vs inference \cite{HEVC, UVG,MCL-JCV} (1080p, 720p, 540p, etc). For position embeddings based on explicit $(x,y)$ pixel coordinates or relative pixel coordinates $(x_i - x_j, y_i -y_j)$, the resultant position sinusoidal harmonics will be very different between training and test (\cref{fig:Mamba bidirectional block}). We found that this caused our models to be over-fit (See Appendix) to the training patch resolution. To combat this, we instead normalize position based on an image token's position within the canonical image plane ($x \in [0,1], y \in [0,1]$) rather than using the raw pixel values for position.

\subsection{Adaptive Latent Skipping}

\begin{table}[h!]
    \centering
    \begin{tabular}{l|cccc}
                       & P Encode & P Decode & B Encode & B Decode\\
                       \midrule
        Latent Skip    & 0.0197s&0.0182s &0.0061s &0.0092s \\
        No Latent Skip &  0.1870s&0.1007s &0.0342s &0.0208s
    \end{tabular}
    \caption{Average entropy coding time (seconds) per 1080p frame for P-frames and B-frames, with and without adaptive latent skipping. Latent skipping reduces entropy coding time by approximately 9× for P-frames and 5× for B-frames.}
    \label{tab:LatentSkip}
\end{table}

We employ an adaptive latent skipping technique to improve runtime efficiency inspired by AlphaVC \cite{shi2022alphavc}. Differing from AlphaVC's fixed threshold approach~\cite{shi2022alphavc}, we introduce a relative thresholding strategy that adapts to each frame's latent distribution:

\begin{align}
    \qlatentat{i} =\begin{cases}
    \mu_i,\sigma_i<\text{mean}(\sigma)\\
    \qlatentat{i}, \sigma_i\geq\text{mean}(\sigma), 
\end{cases}
\end{align}

For multi-stage entropy coding, we compute the threshold separately for each stage using only the $\sigma$ values of elements coded in that stage. When applied to 1080P videos we observe (\cref{tab:LatentSkip}) reductions on the order of $\sim 9\times$ in entropy coding times per frame whilst maintaining comparable performance (see Appendix). The skipping mechanism itself does not introduce asymmetric computation, as the mask M reduces the number of coded symbols symmetrically on both sides.

\subsection{Open GoP and Bidirectional Coding}
Beyond architectural improvements, we apply two inference-time optimizations. 

\textbf{Open GoP:} allows Inter-GoP references: the final B-frame of each GoP can 
reference the next GoP's I-frame \cite{6316136}, providing temporally closer 
references than closed GoP and improving coding efficiency.

\textbf{Bidirectional Coding:} encodes odd-indexed GoPs in both forward and backward directions, selecting the lower rate-distortion result; even-indexed GoPs use closed configuration to maintain linear complexity. While standard in traditional codecs, this is to our knowledge the first application to NVCs. We recommend it for archival or storage-critical scenarios.

\input{figures/combine_bd_rgb.tex}
\subsection{Training}

To train our model, we use a frozen I-frame from DCVC-DC, training both the P-frame model and our new B-frame model with adaptive latent skipping separately.
We use a standard rate distortion loss function\cite{li2023neural} defined as  $L_{RD} =  \lambda D + R$  to train our model. In order to account for the adaptive latent skipping we include a binary mask $M$ in our rate term such that $R= \mathbb{E}_{x\sim p_x} M * -log(p_y)$ where $M$ indicates the latents that were skipped. For our model, we choose to use the same lambda values as \cite{li2023neural} and \cite{li2024neural}, namely $\lambda \in (85,170,380,840)$. Like \cite{li2023neural}, we also have a joint training strategy which first trains our model with separate optimization steps per frame, and then calculates a combined loss function across frames. 

We train our P-frame model for a total of 4.6 million iterations and our B-frame model for 1.85 million iterations. We employ the Adam \cite{kingma2014adam} optimizer with a learning rate of 1e-4 and gradually decrease it to 1e-5. Together, P-frame and B-frame training totals approximately 1.5x the compute of DCVC-DC, with the additional cost attributable to the B-frame fusion module.

%% file: figures/combine_bd_rgb.tex
\begin{table*}[!ht]
    \centering
    
    \adjustbox{width=\textwidth,center}{
    \begin{tabular}{l*{6}{S[table-format=-2.2]@{\,/\,}S[table-format=-2.2]}}
    \toprule
         \textbf{Method} & \multicolumn{2}{c}{\textbf{HEVC B}} & \multicolumn{2}{c}{\textbf{HEVC C}} & \multicolumn{2}{c}{\textbf{HEVC D}} & \multicolumn{2}{c}{\textbf{UVG}} & \multicolumn{2}{c}{\textbf{MCL-JCV}} & \multicolumn{2}{c}{\textbf{Average}} \\
         \midrule
         \multicolumn{13}{l}{\textit{Intra-period = -1 (all frames, single I-frame) / Intra-period = 32 (all frames)}} \\
         \midrule
         VTM-19.0-LDP\cite{VTM} & 0.0 & 0.0 & 0.0 & 0.0 & 0.0 & 0.0 & 0.0 & 0.0 & 0.0 & 0.0 & 0.0 & 0.0 \\ 
        VTM-19.0-RA (Inter-GoP=16) & {\na} & -30.28 & {\na} & -29.07 & {\na} & -24.05 & {\na} & \bfseries -25.87 & {\na} & \bfseries -24.79 & {\na} & -26.82 \\
         DCVC-DC\cite{li2023neural} & -1.48 & -20.93 & -12.58 & -28.63 & -18.41 & -32.87 & 1.09 & -22.90 & -11.61 & -20.67 & -8.59 & -25.20 \\
         DCVC-FM\cite{li2024neural} & -16.95 & -15.90 & -26.38 & -22.60 & -30.16 & -26.28 & -20.67 & -19.06 & -13.21 & -13.40 & -21.47 & -19.44 \\
         DCVC-MB & -29.86 & -28.86 & -35.72 & -33.18 & -44.25 & -39.91 & -23.75 & -21.91 & -15.13 & -13.12 & -29.74 & -27.39 \\
         DCVC-MB-bdc & \bfseries -31.16 & {\na} & \bfseries -36.21 & {\na} & \bfseries -44.25 & {\na} & \bfseries -24.29 & {\na} & \bfseries -16.32 & {\na} & \bfseries -30.45 & {\na} \\
        DCVC-MB-og & {\na} & \bfseries -30.45 & {\na} & -34.07 & {\na} & -40.52 & {\na} & -23.78 & {\na} & -14.30 & {\na} & -28.62 \\
        DCVC-MB-bdc-og & {\na} & -30.38 & {\na} & \bfseries -34.08 & {\na} & \bfseries -40.58 & {\na} & -23.10 & {\na} & -14.99 & {\na} & \bfseries -28.63 \\

         \bottomrule
    \end{tabular}}
    \vspace{0.5em}
    \caption{BD-Rate (\%) in RGB BT.709 (lower is better) for intra-periods of -1 (single I-frame) and 32. Best results are in \textbf{bold}. ``og'' denotes Open GoP and ``bdc'' denotes Bidirectional Coding.}
    \label{tab:combined_bd_rates}
\end{table*}

\begin{table*}[!ht]
    \centering
    
    \begin{tabular}{lcccccc}
    \toprule
         \textbf{Method}&\textbf{HEVC B}  & \textbf{HEVC C}   &  \textbf{HEVC D}  & \textbf{UVG} & \textbf{MCL-JCV} & \textbf{Average}  \\
         \midrule
         VTM-19.0-LDP\cite{VTM} & 0.0 &  0.0 & 0.0 & 0.0 & 0.0 & 0.0  \\
         DCVC-DC\cite{li2023neural} & -19.55 & -20.56 & -33.64 & -26.53 &  -20.29 & -24.11    \\
         DCVC-FM\cite{li2024neural} & -15.78 & -15.56 &  -29.03 & -22.05 & -14.57 & -19.39 \\
         DCVC-LCG\cite{qi2024long} & -17.40 &   -    &   -   & \textbf{-28.51} & \textbf{-20.71} & -22.21 \\
         DCVC-B\cite{sheng2024bidirectionaldeepcontextualvideo}   &   -22.27   &    -19.96    &   -39.29        &         -13.97        & -10.27 & -23.38\\
         \midrule
        %  \midrule
         DCVC-MB & -26.69 & -23.60 & -40.96 & -26.79 & -14.31 & -26.47 \\
         DCVC-MB-og & \textbf{-27.95} & -24.38 & -41.42 & -28.08 & -15.31 & -27.43 \\
         DCVC-MB-bdc-og & -27.31 & \textbf{-24.53} & \textbf{-41.90} & -27.32 & -16.37 & \textbf{-27.49} \\
         \bottomrule
    \end{tabular}
    \vspace{0.5em}
    \caption{BD-Rate (\%) in RGB BT.709 colorspace (lower is better). Intra-period = 32 with 96 frames. Some results are not available due to lack of official implementations. }
    \label{tab:Gop=32,96frames}
\end{table*}

        %  \textbf{Method} & \textbf{HEVC B} & \textbf{HEVC C} & \textbf{HEVC D} & \textbf{UVG} & \textbf{MCL-JCV} & \textbf{Average} \\
        %  \midrule
        %  \multicolumn{7}{l}{\textit{Intra-period = -1 (all frames, single I-frame)}} \\
        %  \midrule
        %  VTM-19.0-LDP & 0.0 & 0.0 & 0.0 & 0.0 & 0.0 & 0.0 \\ 
        % VTM-19.0-RA (inter-gop=16) & $\emptyset/-30.28$ & $\emptyset/$-29.07 & $\emptyset/$-24.05 & \textbf{$\emptyset/$-25.87} & \textbf{$\emptyset/$-24.79} & $\emptyset/$-26.82 \\
        %  DCVC-DC & -1.48/-20.93 & -12.58/-28.63 & -18.41/ -32.87 & 1.09/ -22.90 & -11.61/-20.67 & -8.59/-25.20 \\
        %  DCVC-FM & -16.95/-15.90 & -26.38/-22.60 & -30.16/-26.28 & -20.67/-19.06 & -13.21/-13.40 & -21.47/-19.44 \\
        %  DCVC-MB & -29.86/-28.86 & -35.72/-33.18 & -44.25/-39.91 & -23.75/-21.91 & -15.13/-13.12 & -29.74/-27.39 \\
        %  \textbf{DCVC-MB-bdc} & \textbf{-31.16}/$\emptyset$ & \textbf{-36.21}/$\emptyset$ & \textbf{-44.25}/$\emptyset$ & \textbf{-24.29}/$\emptyset$ & \textbf{-16.32}/$\emptyset$ & \textbf{-30.45}/$\emptyset$ \\
        % DCVC-MB-og & $\emptyset$/\textbf{-30.45} & $\emptyset$/-34.07 & $\emptyset$/-40.52 & $\emptyset$/-23.78 & $\emptyset$/-14.30 & $\emptyset$/-28.62 \\
        %  \textbf{DCVC-MB-bdc-og} & $\emptyset$/-30.38 & \textbf{$\emptyset$/-34.08} & \textbf{$\emptyset$/-40.58} & $\emptyset$/-23.10 & $\emptyset$/-14.99 & \textbf{$\emptyset$/-28.63} \\

%% file: sections/4_experiments.tex
\section{Experiments}
\subsection{Experimental Details}
\textbf{Dataset:} We train our model on the train split of the Vimeo90k dataset \cite{Vimeo90k}, using 7-frames with a patch resolution of $256 \times 256$. For validation, we use BVI-DVC\cite{BVI-DVC}.
\newline\textbf{Test Conditions:} 
We evaluate our model on the HEVC \cite{HEVC}, UVG \cite{UVG}, and MCL-JCV \cite{MCL-JCV} datasets. To convert these originally YUV-based datasets to RGB, we use the BT.709 full-range conversion matrix. We compare against VTM-19.0-LDP/RA \cite{VTM}, as well as DCVC-DC \cite{li2023neural} and DCVC-FM \cite{li2024neural} as our SOTA NVCs. We choose to omit other neural video codecs using B-frames as prior works often fall short of even P-frame based NVC's. For comparison, we use two major test conditions: Intra-period = 32 with all frames (Table \ref{tab:combined_bd_rates}), and Intra-period = -1 with all frames (Table \ref{tab:combined_bd_rates}). This allows us to compare directly with previous works \cite{li2022hybrid,li2023neural,li2024neural,ho2022canf} using sequences in their entirety. Performance is measured using BD-Rate \cite{bjontegaard2001calculation} with RGB PSNR. Despite not optimizing for YUV420 we use a naive RGB to YUV conversion and measure the BD-Rate in the Appendix.

\begin{figure*}[!h]
    \centering
    \includegraphics[width=\linewidth]{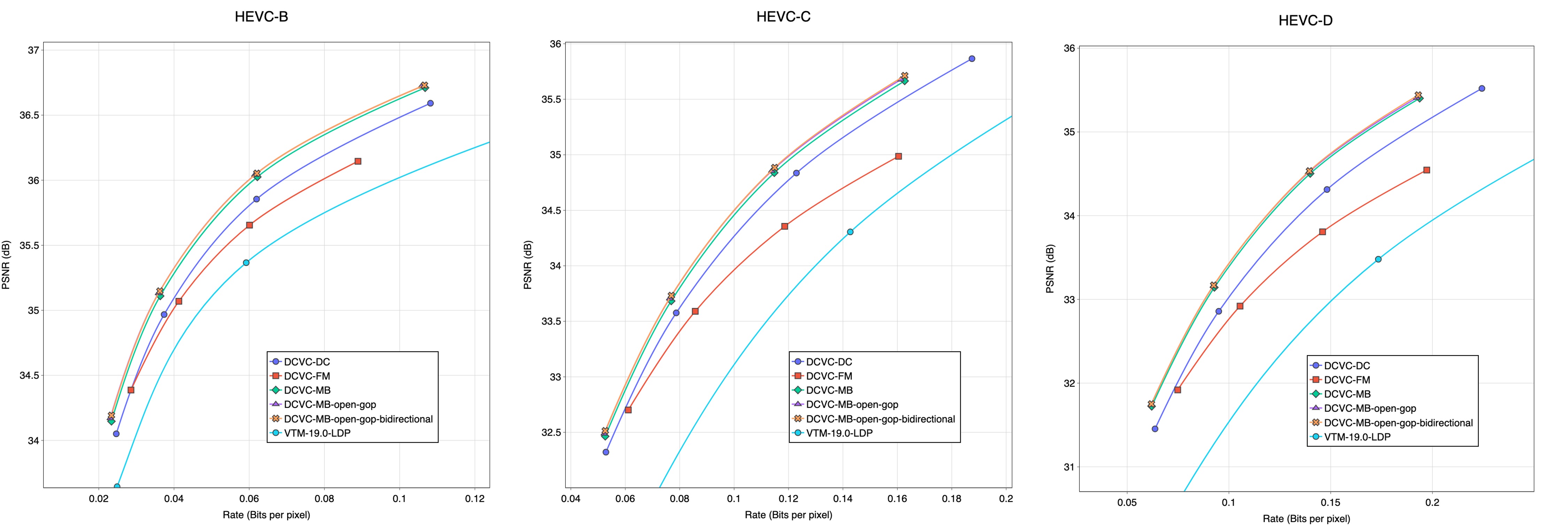}
    \caption{RD-Curves for Intra-period=32, all frames for HEVC-B, HEVC-C, and HEVC-D. Additional curves in the Appendix.}
    \label{fig:rd_curves}
    \vspace{-5mm}
\end{figure*}

\subsection{Comparisons with Previous SOTA Methods}

\begin{figure*}[!t]
    \centering
    \includegraphics[width=.9\textwidth]{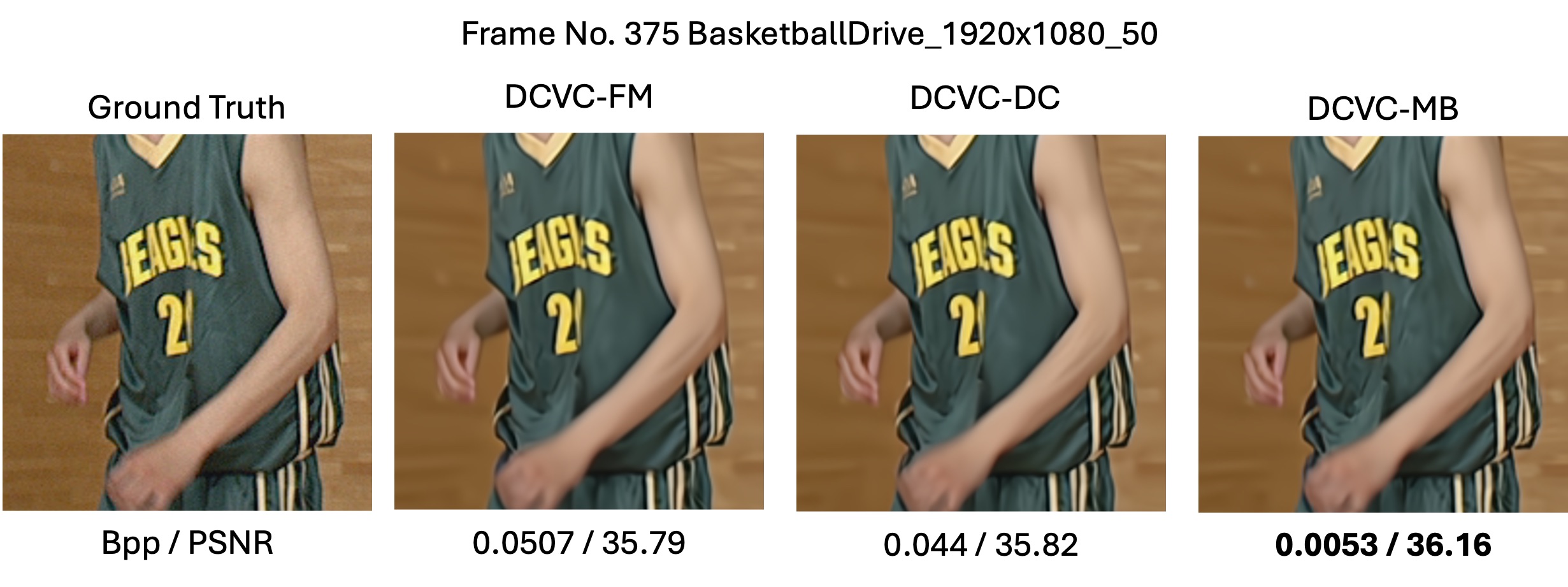}
    \caption{Subjective quality comparison of DCVC-MB versus DCVC-DC and DCVC-FM. Note the reconstruction of the wrinkles in the player's jersey which is accomplished with an order of magnitude less Bpp. Additional comparisons in the Appendix.}
    \label{fig:subjective_comp}
\end{figure*}
Our results (Table \ref{tab:combined_bd_rates}) show that DCVC-MB outperforms VTM, DCVC-DC, and DCVC-FM on average across all datasets (\cref{fig:rd_curves}). We report three settings: the base model, open-GoP (og), and open-GoP with bidirectional coding (bdc-og). For Intra-period = -1, we report only the first two, as open-GoP is redundant with a single GoP.

Beyond the BD-rate improvements over VTM-19.0-LDP shown in our tables, we achieve average BD-Rate reductions of \textbf{3.43\%} (Intra-period=32) and \textbf{21.86\%} (Intra-period=-1) over DCVC-DC, and \textbf{8.21\%} and \textbf{8.98\%} over DCVC-FM in the same settings. Full hierarchical VTM-RA (GoP=32) yields only marginal additional BD-rate savings over Inter-GoP=16\cite{JVET-T0063}, attributable to per-layer QP offsets incompatible with our IBP structure; we therefore compare against VTM-RA with Inter-GoP=16. DCVC-MB is shown (Table \ref{tab:combined_bd_rates}) to be the only NVC that surpasses the stronger VTM configuration and does so by $-1.81\%$. We would like to note that DCVC-LCG\cite{qi2024long} does not have a public implementation, so we report their results as-is. DCVC-B has a hard sequence length limitation, so we report only 96-frame results; DCVC-MB outperforms both by significant margins (Table \ref{tab:Gop=32,96frames}). Subjectively, DCVC-MB preserves finer image details while using significantly fewer bits compared to both DCVC-DC and DCVC-FM (\cref{fig:subjective_comp})

%% file: sections/ablation.tex
\section{Ablation Study} 
\begin{table}[th!]
    \centering
    \caption{Ablation Studies on HEVC-C vs DCVC-DC.}
    \label{tab:ablation}
    \resizebox{0.5\textwidth}{!}{
            \begin{NiceTabular}[t]{|c|c|c|c|c|c|c|c|}
            \CodeBefore
                %   \columncolor{gray!15}{2,4,6,8}
                \Body
                \toprule
                 & $M_a$ & $M_b$ & $M_c$ & $M_d$ & $M_e$  & $M_f$ & $M_g$ \\
                \midrule
                \makecell{CNN Fusion} &  & \checkmark  &   &  \\
                \midrule
                \makecell{Mamba Fusion }  &  & & \checkmark& \checkmark & \checkmark  & \checkmark  & \checkmark \\
                \midrule
                \makecell{Positional Embedding} &   & & & \checkmark&      &   & \\
                \midrule
                \makecell{Canonical Positional \\Embedding} && & && \checkmark  & \checkmark  & \checkmark \\
                \midrule
                \makecell{Open GoP}  &&&&  &  & \checkmark & \checkmark \\
                \midrule
                \makecell{Bidirectional \\ Coding}  & &&&& &  &  \checkmark \\
                \midrule
                BD-Rate(\%) & 0.0 & -2.5 & -3.03  & 13.38 & -4.55  & -5.44 & -5.45\\
                \bottomrule
            \end{NiceTabular}
        }
\end{table}

To validate the effectiveness of each component in our system, we present an ablation study in Table~\ref{tab:ablation}. We use DCVC-DC as the baseline model $M_a$, which has no fusion architecture. The first modification adds a CNN fusion architecture resulting in model $M_b$. While $M_b$ improves performance over $M_a$, we observe further gains by adopting our Mamba-based fusion model $M_c$. Despite $M_c$ requiring an additional forward pass, it consistently achieves superior BD-Rate performance across all test conditions. Next, we evaluate positional encoding strategies. Model $M_d$, which uses standard 2D positional embeddings, fails to generalize to unseen resolutions (See Appendix). Replacing this with 2D canonical positional embeddings in model $M_e$ yields further performance improvements over $M_c$ (\cref{tab:ablation}).Finally we test the inference-time strategies. Model $M_f$ incorporates open GoP, achieving a BD-Rate savings of approximately 0.90\% over $M_e$, with no additional encoding or decoding overhead. Finally, model $M_g$ applies Bidirectional Coding, which requires encoding every other GoP twice, doubling encoding time. Though the BD-Rate gains are modest compared to open GoP alone, they are measurable. In scenarios where open GoP is not applicable, Bidirectional Coding still provides a 0.91\% BD-Rate improvement over $M_c$ (Table~\ref{tab:combined_bd_rates}). The base Mamba fusion model accounts for the majority of BD-Rate gains $(-4.55\%)$, with inference-time strategies open GoP ($M_f$) and bidirectional coding ($M_g$) providing additive but secondary improvements ($-0.89\%$ and $-0.01\%$, respectively).

%% file: sections/5_conclusion.tex
\section{Conclusion}
We present DCVC-MB, a novel neural B-frame video codec that leverages bidirectional Mamba fusion and adaptive latent skipping to achieve state-of-the-art compression performance. Our method demonstrates notable BD-rate improvements over both traditional codecs (VTM-19.0) and leading neural codecs (DCVC-DC, DCVC-FM), with gains of up to 30.45\% and 28.63\% in different test configurations. The bidirectional Mamba architecture effectively exploits temporal redundancy while maintaining computational efficiency through our adaptive skipping mechanism. These results establish that DCVC-MB is a competitive B-frame neural video codec and represents progress toward practical neural video compression.

%% file: sections/x_appendix.tex
% % \clearpage
% % \setcounter{page}{1}

\section{Memory Scaling}
\begin{figure}[h!]
    \centering
    \includegraphics[width=\linewidth]{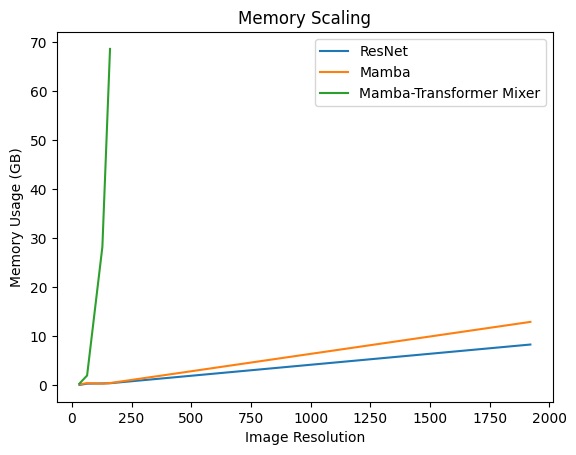}
    \caption{Comparing the inference memory cost of DCVC-MB at varying resolutions for Mamba, CNN, and Transformer based fusion. The Transformer architecture is an encoder-only model, using nheads=8 and same $d\_model$ and $L$ layers as the Mamba and CNN architecture. We use the latest PyTorch-enabled flash-attention to ensure a fair comparison.}
    \label{fig:VRAMScaling}
\end{figure}

Transformer models are known for their memory and compute scaling requirements w.r.t to sequence length (i.e. $O(N^2)$). For our fusion model design, this quickly results in Out of Memory errors. Take, for example, our $C^0$ fusion model, which fuses inputs $C^{0}_{t+1}$ and $C^0_{t-1}$ $\in  \mathbb{R}^{(B,48,H,W)}$. After CNN embedding, this results in an $\textbf{x} \in \mathbb{R}^{(B,M,48)}$ where $M= H \times W$. This results in 65536 tokens at an input resolution of $(256 \times 256)$. Given self-attention scaling, Figure \ref{fig:VRAMScaling}, this fusion model design quickly becomes untenable to train with currently existing hardware.

\section{Runtime Complexity Analysis}
\begin{table}[!h]
\centering
\small
\setlength{\tabcolsep}{2pt}
\begin{tabular}{l|cc|cc|c}
\toprule
 & \multicolumn{2}{c|}{Enc (ms/frame)} & \multicolumn{2}{c|}{Dec (ms/frame)} & Memory \\
 & P & B & P & B & (GB) \\
\midrule
Ours (skip) & 323 & 626 & 141 & 492 & 13.8 \\
\bottomrule
\end{tabular}
\caption{Runtime complexity and memory analysis.}
\label{tab:runtime}
\end{table}

The B-frame model is approximately 2× (Table \ref{tab:runtime}) the latency of the P-frame (same as DCVC-DC\cite{li2023neural}) model. Crucially, our
Mamba-based fusion scales linearly O(N) with spatial resolution, whereas a transformer-based alternative would scale quadratically O(N²). As shown in Figure \ref{fig:VRAMScaling}, a
transformer fusion model exceeds 40GB VRAM at 1080p,
while our full B-frame pipeline fits within 13.8GB.

\section{YUV to RGB Conversion}

To convert our YUV test sequences to RGB, we use a full-range BT.709 based matrix. We use the following format for the ffmpeg conversion: 

\noindent\texttt{ffmpeg -f rawvideo -vcodec rawvideo -s 416x240 -r 25 -pix\_fmt yuv420p  -i [input.yuv] -pix\_fmt rgb24}\texttt{-vf scale\=in\_range=full:in\_color\_matrix=bt709}
\texttt{:out\_range=full:out\_color\_matrix=bt709} 
\texttt{-color\_primaries bt709 -color\_trc bt709 -colorspace bt709 -y [output\_rgb\_\%03.png]}

\section{IBP Encoding Algorithm}

Here we provide an explicit approach for IBP encoding process: \cref{alg:cap} for the closed GoP case and \cref{alg:oap} for the open GoP case. Note that \cref{alg:cap} is per GoP while \cref{alg:oap} describes how to encode an entire sequence.

\begin{algorithm}[!h]
\caption{IBP Encoding Closed GoP}\label{alg:cap}
\begin{algorithmic}[1]
\State $X \gets [x_1,x_2,...,x_n] $
\State $i \gets 0$
\State $N \gets \text{len}(X)$
\State $dpb \gets []$
\While{$i < N$ }
    \If{$i = 0$}
        \State $dpb[i] \gets IFrame(X_i)$ 
    \ElsIf{$i = N-1$}
        \State $dpb[i] \gets BFrame(X_i, dpb[i-3], dpb[i-1])$
    \ElsIf{$i \text{ is odd}$}
        \State $dpb[i+1] \gets PFrame(X_{i+1}, dpb[i-1])$
    \ElsIf{$i \text{ is even}$}
        \State $dpb[i-1] \gets BFrame(X_{i-1}, dpb[i-2], dpb[i+1])$
    \EndIf
    \State $i \gets i+1$
\EndWhile
\end{algorithmic}
\end{algorithm}

\begin{algorithm}[]
\caption{IBP Encoding Open GoP}\label{alg:oap}
\begin{algorithmic}[1]
\State $X \gets [x_1,x_2,...,x_t] $
\State $i \gets 0$
\State $T \gets \text{len}(X)$
\State $N \gets \text{gop\_size}$
\State $dpb \gets []$
\While{$i < T$}
    \If{$i = 0 \text{ or } ((i+1) \bmod N = 0 \text{ and } (i < T-1))$}  
        \State $dpb[i] \gets IFrame(X_i)$ 
    \ElsIf{$i = T-1$}
        \State $dpb[i] \gets BFrame(X_i,dpb[i-3],dpb[i-1])$
    \ElsIf{$i \text{ is odd}$}
        \State $dpb[i+1] \gets PFrame(X_{i+1},dpb[i-1])$
    \ElsIf{$i \text{ is even}$}
        \State $dpb[i-1] \gets BFrame(X_{i-1},dpb[i-2],dpb[i+1])$
    \EndIf
    \State $i \gets i+1$
\EndWhile
\end{algorithmic}
\end{algorithm}

\begin{figure}[!t]
    \centering
    \includegraphics[width=.9\linewidth]{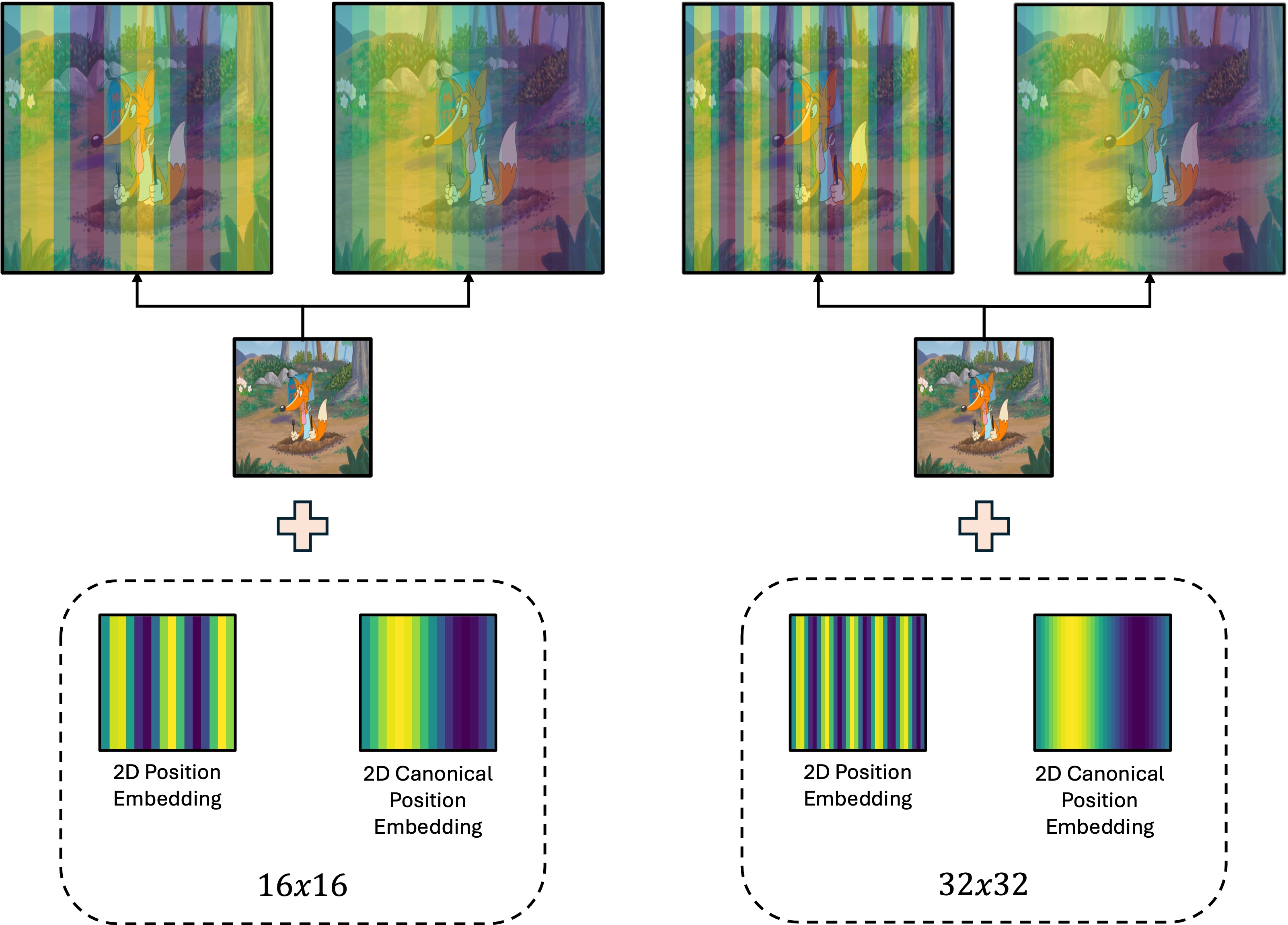}
    \caption{Diagram comparing the $0^{th}$ sinusoidal harmonic of 2D canonical position embedding vs 2D resolution dependent position embedding at varying resolutions.}
    \label{fig:CanonicalPositionEmbedding}
\end{figure}

\section{Canonical Position Embedding} 
\label{Canonical Position Emb}
\begin{table}[!h]
  \centering
  \begin{tabular}{ccccc}
    \toprule
    \textbf{ BD-Rate } & DCVC-DC & $\textbf{p}_{\textbf{s}}$ & no $\textbf{p}_{\textbf{s}}$ & $\textbf{p}_{\textbf{ce}}$ \\
    \midrule
    Class B & 0.0 & 55.12  & -7.34  & \textbf{-7.93}  \\
    Class C & 0.0 &  14.27 & -3.03  & \textbf{-4.55}  \\
    Class D & 0.0 & -6.96 & \textbf{-7.14} & -7.04 \\
    \bottomrule
  \end{tabular}
  \vspace{1em}
  \caption{Comparing results of DCVC-MB using different position embeddings vs DCVC-DC. BT709 HEVC Dataset, Intra-period = 32 All frames.}
  \label{tab:CanonicalPositionEmbeddingTable}
\end{table}
For canonical position embedding, denoted as $\mathbf{p}_{\textbf{ce}}$, we adopt the same 2D sinusoidal basis functions as described in prior works on positional encoding \cite{Conditional2DPositionEncoding,Explicit2DPositionEncoding} (see \cref{eq:CanonicalPositionEmbeddingEq}). The key difference lies in how the input coordinates are handled: instead of using absolute pixel indices, we normalize spatial coordinates to a canonical image plane, where $x, y \in [0, 1]$.

This normalization ensures that the resulting positional encoding harmonics are resolution-agnostic (i.e., they remain consistent across input sizes) modulo the sampling frequency. As a result, higher-resolution inputs produce interpolated harmonics rather than extrapolated ones, improving the generalization of the model to unseen resolutions(\cref{fig:CanonicalPositionEmbedding}).

As shown in Table~\ref{tab:CanonicalPositionEmbeddingTable}, the canonical positional embedding $\mathbf{p}_{\textbf{ce}}$ outperforms both standard 2D positional embedding $\mathbf{p}{_\textbf{s}}$ and models without any positional encoding. Notably, $\mathbf{p}_{\textbf{s}}$ exhibits poor generalization, performing best on resolution classes closest to the training resolution (e.g., Class D at $416 \times 240$) and worst on larger resolutions such as Class B.

\begin{equation}
\label{eq:CanonicalPositionEmbeddingEq}
\resizebox{1\hsize}{!}{
$
    \textbf{p}_{\textbf{ce}}(i,j,k) =  \\
\begin{cases}
     & sin(\frac{2\pi i}{W} \times e^{\frac{4klog(10,000)}{L}}), \text{if } k  \text{ is even } \text{ and } (k < \frac{L}{2}) \\
     & cos(\frac{2\pi i}{W} \times e^{\frac{4klog(10,000)}{L}}), \text{if } k  \text{ is odd  } \text{ and } (k < \frac{L}{2}) \\ 
     & sin(\frac{2\pi j}{H} \times e^{\frac{4klog(10,000)}{L}}), \text{if } k  \text{ is even } \text{ and } (k \geq \frac{L}{2}) \\
    & cos(\frac{2\pi j}{H} \times e^{\frac{4klog(10,000)}{L}}), \text{if } k  \text{ is odd  } \text{ and } (k \geq \frac{L}{2})  
\end{cases}
\\  \textrm{ s.t. } \begin{matrix}
 i \in [0, W - 1]
 \\j \in [0, H - 1]
 \\k \in [0, L - 1] &\\ 
 \end{matrix}
$
}
\end{equation}

\begin{figure*}[bh!]
  \centering
  \includegraphics[width=\linewidth]{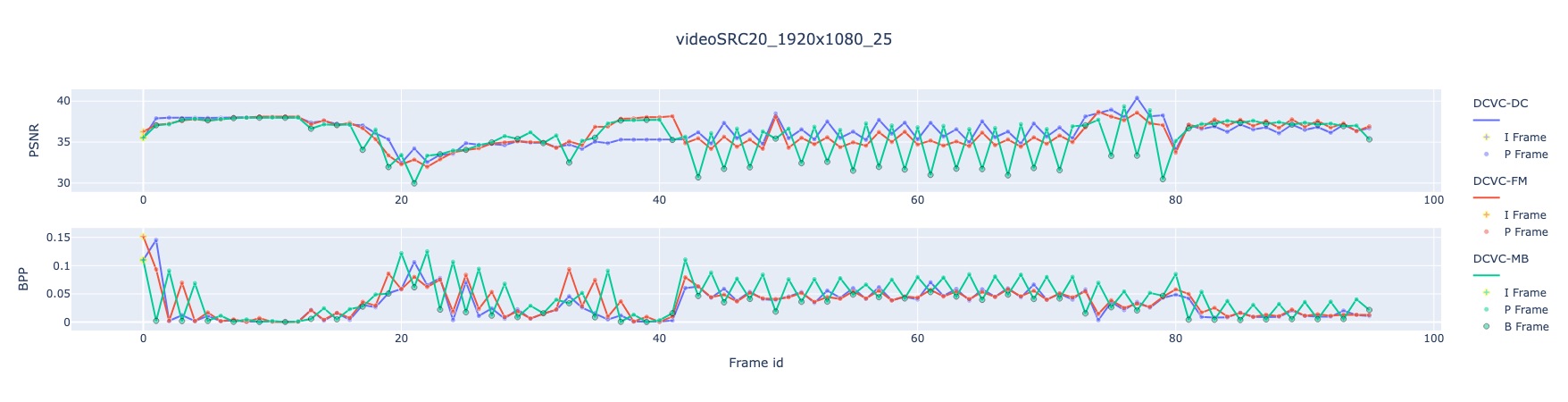}
   \caption{Inspecting the frame by frame PSNR vs BPP we see that there are cases where our B-frame model can fail to incorporate the information from the reference frames. This is commonly seen on animated content which is drastically different from the data contained in Vimeo90k as the subjects and motion between frames in animated content is drastically different from natural video.}
   \label{fig:mcl-jcv}
   \setlength{\belowcaptionskip}{-10pt}
\end{figure*}

\section{Adaptive Latent Skipping:}
\begin{table}[!h]
  \centering
  % \caption{BD-Rate (\%) comparison in RGB colorspace measured with PSNR. The anchor is VTM-17.0.}
   % \renewcommand{\arraystretch}{1.2}
    \small

        \begin{tabular}{lccc}
            \toprule
\textbf{Method} & HEVC-B & HEVC-C& HEVC-D\\  \hline
DCVC-DC & 0 & 0 & 0 \\
Adaptive Latent Skipping & –3.15 & 0.59 & -0.98\\
No Skip   & 3.76 & 4.88 & 1.98 \\
AlphaVC\cite{shi2022alphavc}   & 8.08 & 12.37 & 10.72\\

% AdaSkip –3.15 0.59 –0.98
% No Skip 3.76 4.88 1.98
% AlphaVC-Thresh 8.08 12.37 10.72
\bottomrule
    \end{tabular}
    \vspace{1em}
    \caption{We test the hard threshold vs our adaptive latent skipping method and use DCVC-DC results as the anchor. We observe that the hard threshold strategy results in degraded performance in terms of BD-Rate.}
  \label{table:abl_study}
\end{table}
 In order to verify the effectiveness of our adaptive latent skipping, we compare against the hard threshold proposed in AlphaVC \cite{shi2022alphavc}. We limit this experiment to just the DCVC-DC P frame model which is the same as our main model.
 We test AlphaVC's method with a $95\%$ confidence and train DCVC-DC with AlphaVC latent skipping and evaluate on HEVC classes B-D. Table \ref{table:abl_study} shows that the hard threshold often results degraded performance over the vanilla DCVC-DC model. In contrast, our adaptive latent skipping is able to improve upon the model for classes B and D which verifies the effectiveness. Additionally, when a model is trained with adaptive latent skipping but does perform latent skipping at inference time we observe a noticeable performance drop. This suggests that the model may be biased towards the slightly noisier latent elements introduced from skipping.    

 \begin{table}[h]
\centering
\small
\setlength{\tabcolsep}{4pt}
\begin{tabular}{l|cccc}
\toprule
Rate Index ($\lambda$) & 0 (low) & 1 & 2 & 3 (high) \\
\midrule
P-frame skip ratio & 0.74 & 0.74 & 0.71 & 0.68 \\
B-frame skip ratio & 0.53 & 0.62 & 0.75 & 0.85 \\
\bottomrule
\end{tabular}
\caption{Skip ratios for P-frames and B-frames at different rate indices.}
\label{table:skip_ratio}
\end{table}

From Table \ref{table:skip_ratio} we can observe that as the quality increases
the skip ratio for P-frames decreases. Conversely, the B-
frame actually increases the amount of skipped latents. This
demonstrates the effectiveness of our B-frame fusion model
as we are more effectively utilizing information from the
P-frames. The decoder processes a fixed-size tensor regard-
less of skipping.

\section{Limitations}
\subsection{Naive YUV comparisons}

\begin{table}[!h]
    \centering
    \begin{tabular}{lcccc}
        \toprule
        \textbf{YUV420 Intra-period=32} & \textbf{HEVC-B} & \textbf{HEVC-C} & \textbf{HEVC-D} & \textbf{Average} \\
        \midrule
        VTM-RA (Intergop=16) & 0 & 0 & 0 & 0 \\
        DCVC-DC & 18.72 & -0.29 & -10.08 & 2.78 \\
        DCVC-FM & \textbf{12.72} & \textbf{-5.92} & -15.24 & \textbf{-2.81} \\
        DCVC-MB (ours) & 16.03 & -1.65 & \textbf{-17.32} & -0.98 \\
        \bottomrule
    \end{tabular}
    \vspace{1em}
    \caption{BD-Rate (\%) comparison in YUV420 colorspace with Intra-period = 32.}
    \label{tab:combined_bd_rates_yuv420}
\end{table}

We do acknowledge that YUV420 is an important color space to test and include a naive comparison by converting our RGB model outputs back into YUV420 for comparison. However, we found that training a YUV model was much more difficult than RGB due to the challenges of balancing the distortion across the three channels. Despite not optimizing for YUV420 directly, we still see strong performance against VTM-19.0 and previous SOTA NVCs in \cref{tab:combined_bd_rates_yuv420}.

% Even though we lack a proper training recipe for DCVC-MB on YUV420, we provide a first order approximation using a RGB-to-YUV420 conversion to compare the DCVC-MB outputs with the results of previous works and VTM-19.0

\subsection{Domain Gap}
While our approach has stronger performance on most datasets, we realize that there are cases where our model has suboptimal performance. For example, our performance on MCL-JCV on the intra-period=32 setting is worse than one of our baselines, DCVC-DC. After close analysis, we find that our B-frame model performance is our most common failure case. 
In the following graph, we observe that on videoSRC20 \cref{fig:mcl-jcv} of the MCL-JCV dataset, we see that our B-frame model has lower PSNR and higher BPP than DCVC-DC. 
This is likely due to the domain gap from the training data. Specifically, when looking ad videoSRC20 in the regions where we have higher distortion we can see there is significantly larger motion between each frame than what you would see in natural videos. As a result our B-frame model is not trained to properly handle such drastic changes between neighboring frames as well as the subject matter.

\section{RD-Curves}
\begin{figure*}[!htbp]
    \centering
    \includegraphics[width=\textwidth]{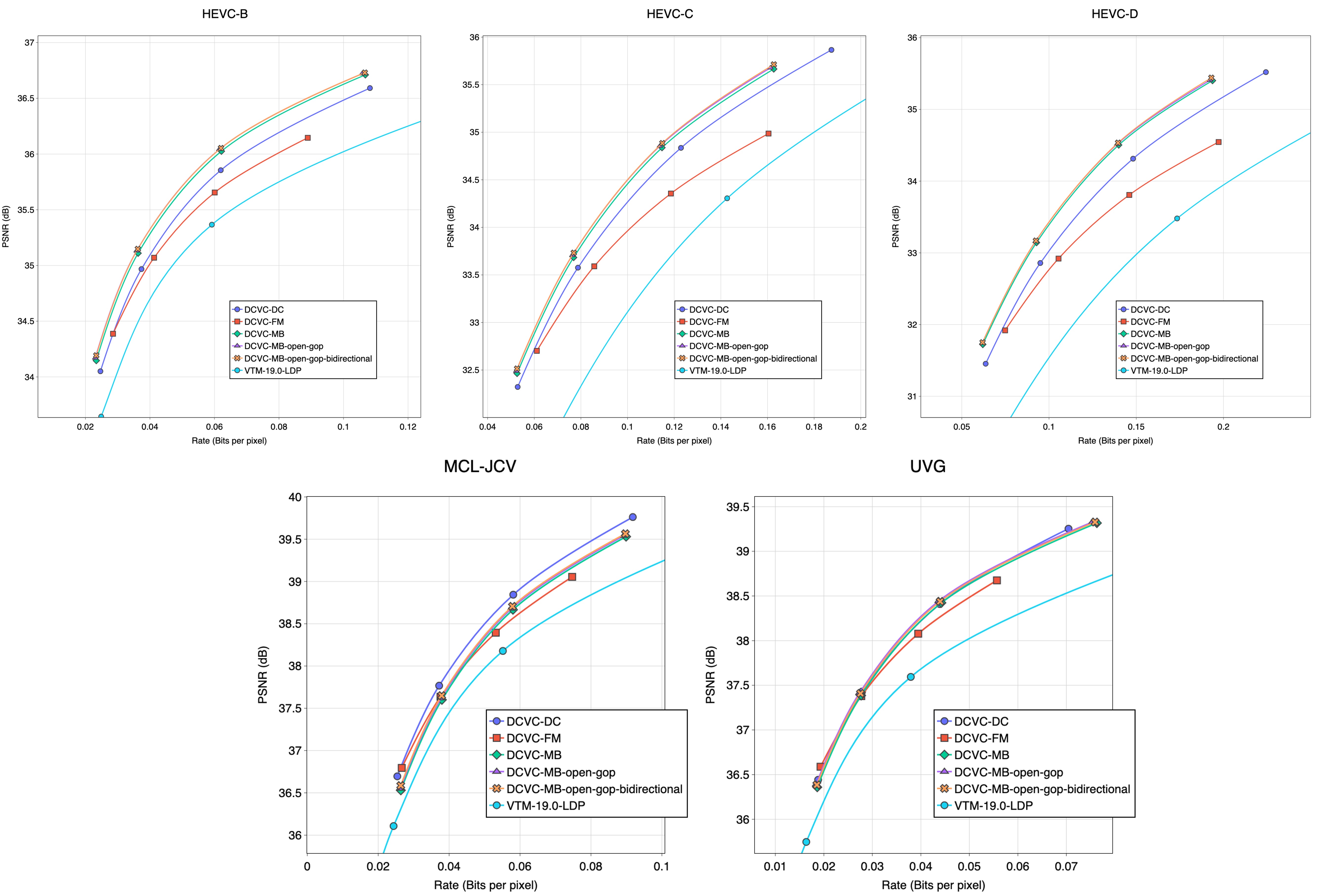}
    \caption{RD-Curves for intra-period = 32, All frames}
    \label{fig:AllFramesRDCurveGoP32}
\end{figure*}

\begin{figure*}[!htbp]
    \centering
    \includegraphics[width=\textwidth]{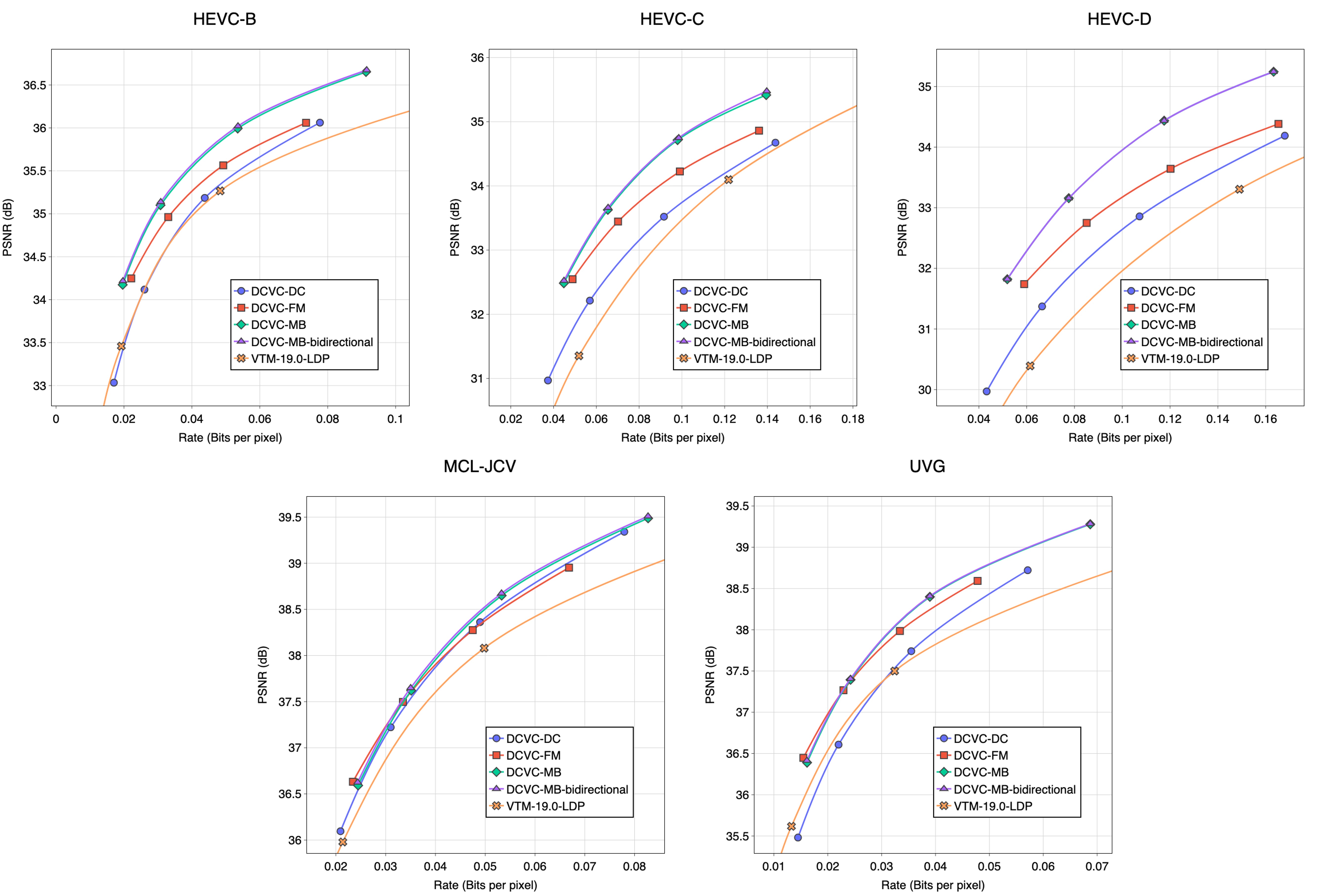}
    \caption{RD-Curves for intra-period = -1, All frames}
    \label{fig:AllFramesRDCurveGoP9999}
\end{figure*}

\begin{figure*}[!htbp]
    \centering
    \includegraphics[width=\textwidth]{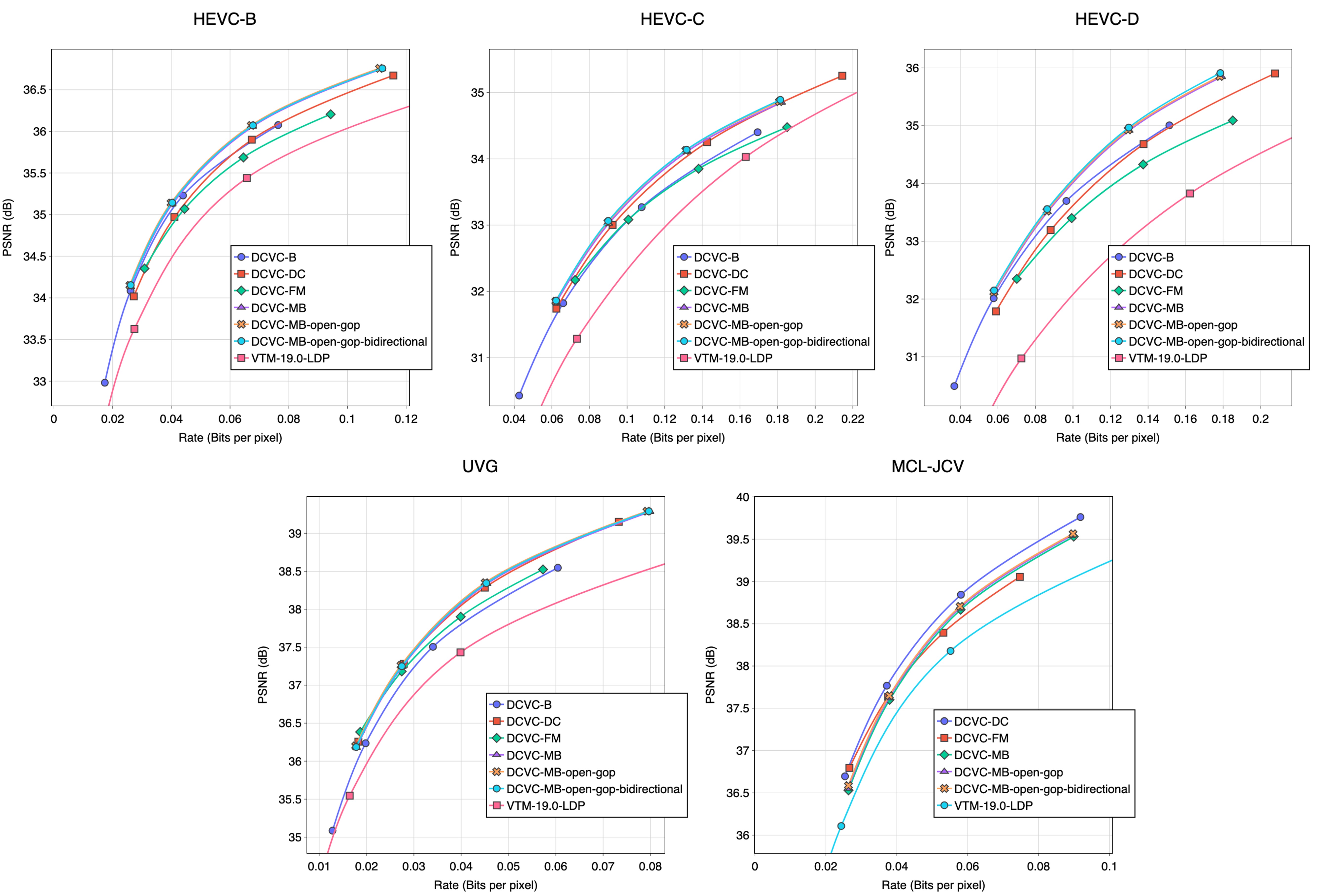}
    \caption{RD-Curves for intra-period = 32, 96 frames}
    \label{fig:96FramesRDCurveGoP32}
\end{figure*}

\begin{figure*}[!htbp]
    \centering
    \includegraphics[width=\textwidth]{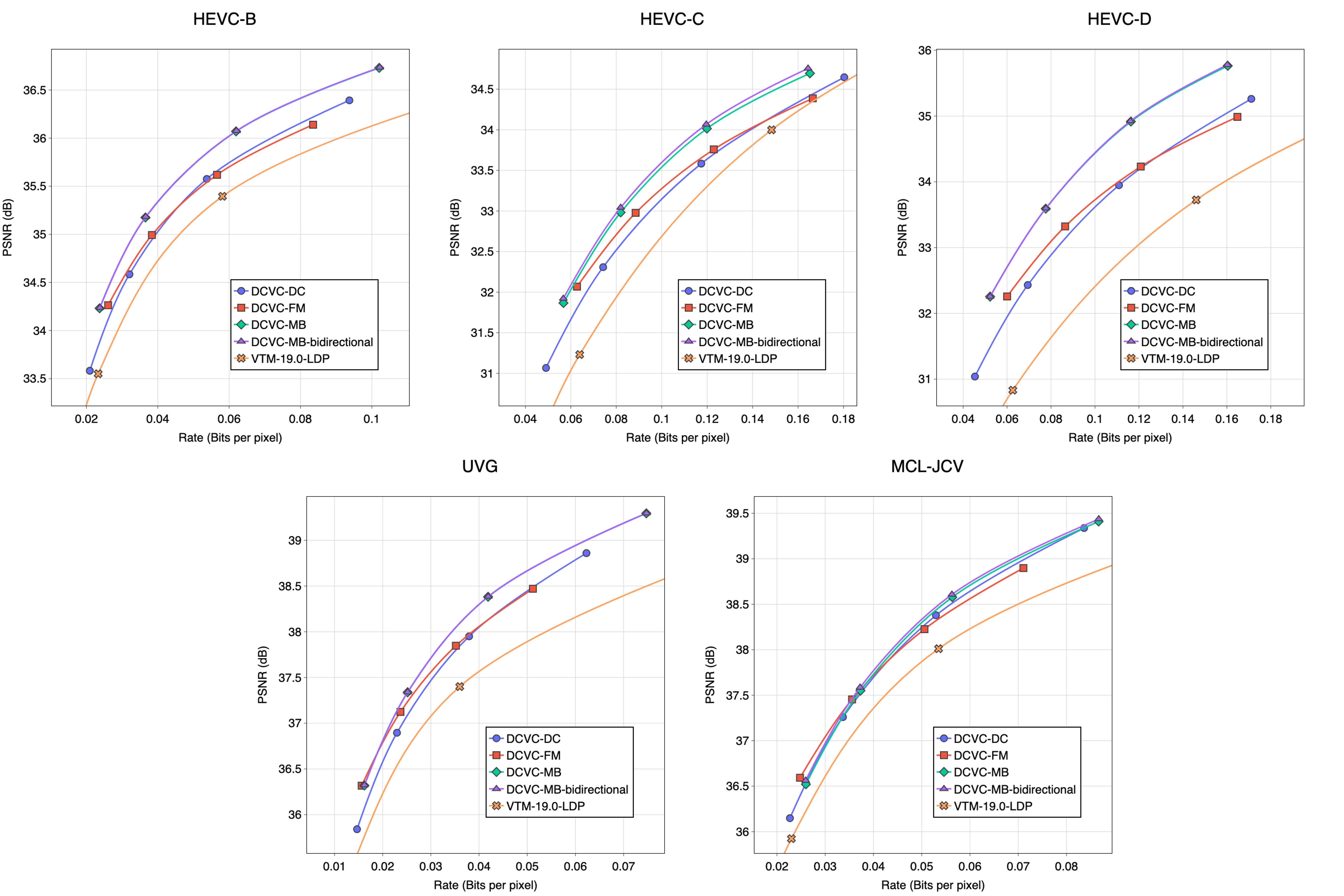}
    \caption{RD-Curves for intra-period = -1, 96 frames}
    \label{fig:96FramesRDCurveGoP9999}
\end{figure*}

Here we include the RD-Curves for all test conditions discussed: All frames intra-period = 32 (\cref{fig:AllFramesRDCurveGoP32}), 
All frames intra-period = -1 (\cref{fig:AllFramesRDCurveGoP9999}), 96 frames intra-period = 32 (\cref{fig:96FramesRDCurveGoP32}), and 96 frames intra-period = -1 (\cref{fig:96FramesRDCurveGoP9999}). 

\section{Visual Comparisons}
\begin{figure*}[!h]
    \centering
    \includegraphics[scale=.15]{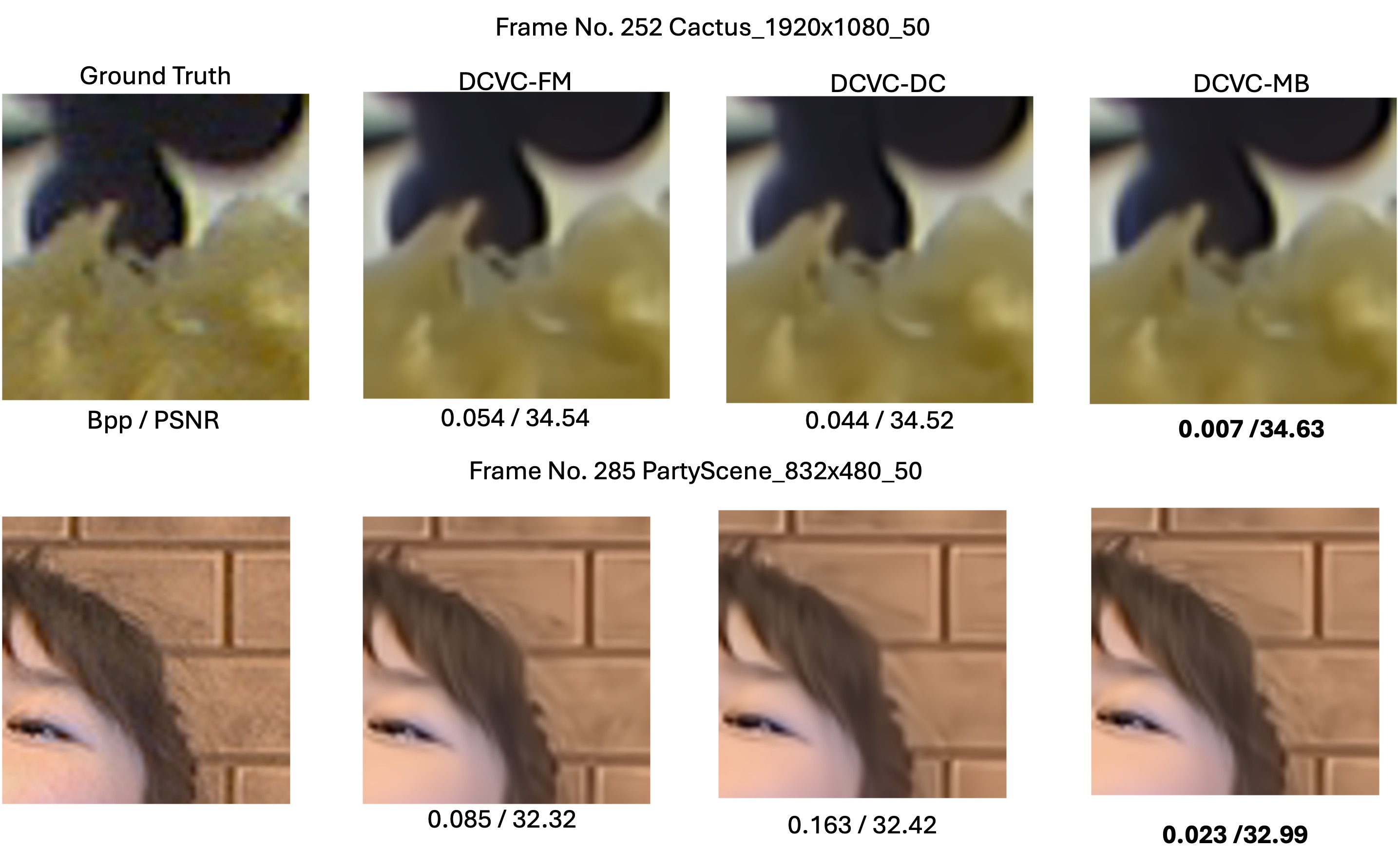}
    \caption{In this example DCVC-MB is able to maintain a higher PSNR and preserve the details where the flower meets the playing card while maintaining an order of magnitude lower bpp as well as more accurately preserve the texture of the bricks on the wall while maintaining a lower bpp.}
    \label{fig:subjective}
\end{figure*}
% \begin{figure*}[!hbp]
%     \centering
%     \includegraphics[scale=.50]{figures/cactus_example.png}
%     \caption{In this example DCVC-MB is able to maintain a higher PSNR and preserve the details where the flower meets the playing card while maintaining an order of magnitude lower bpp.}
%     \label{fig:cactus}
% \end{figure*}
We provide additional qualitative examples and compare against both DCVC-DC\cite{li2023neural} and DCVC-FM\cite{li2024neural}. In order to showcase the advantages of DCVC-MB we showcase frames that are coded as B-frames in \cref{fig:subjective}.
\clearpage